\documentclass[review]{elsarticle}

\usepackage{lineno,hyperref}
\modulolinenumbers[5]
\usepackage[english]{babel}
\usepackage{tikz}
\usetikzlibrary{calc}
\journal{Journal of \LaTeX\ Templates}
\usepackage{amsmath,amssymb,amsthm}
\usepackage{lineno,hyperref,verbatim}
\modulolinenumbers[5]
\usepackage{adjustbox}
\usepackage{multirow}
\usepackage{algpseudocode}
\usepackage{algorithm}

\newcommand{\Beq}{\begin{equation}}
\newcommand{\Eeq}{\end{equation}}
\newcommand{\beq}{\begin{equation*}}
\newcommand{\eeq}{\end{equation*}}
\newcommand{\bal}{\begin{align}}
\newcommand{\eal}{\end{align}}

\newcommand{\g}{\gamma}

\newcommand{\bpr}{\begin{proof}}
	\newcommand{\epr}{\end{proof}}

\newcommand{\bel}[1]{\begin{equation}\label{#1}}
\newcommand{\ee}{\end{equation}}

\usepackage[normalem]{ulem}

\theoremstyle{definition}

\begin{document}

\begin{frontmatter}

\title{A method for large diffeomorphic registration via broken geodesics}


\author[IIIT]{Alphin J Thottupattu\corref{correspondingauthor}}
\cortext[correspondingauthor]{Corresponding author}
\ead{alphinj.thottupattu@research.iiit.ac.in}

\author[IIIT]{Jayanthi Sivaswamy}
\author[TIFAR]{Venkateswaran P.Krishnan}

\address[IIIT]{International Institute of Information Technology, Hyderabad 500032,
India}
\address[TIFAR]{TIFR Centre for Applicable Mathematics, Bangalore 560065, India}

\begin{abstract}
Anatomical variabilities seen in longitudinal data or inter-subject data is usually described by the underlying deformation, captured by non-rigid registration of these images. Stationary Velocity Field (SVF) based non-rigid registration algorithms are widely used for registration. SVF based methods form a metric-free framework which captures a finite dimensional submanifold of deformations embedded in the infinite dimensional smooth manifold of diffeomorphisms. However, these methods cover only a limited degree of deformations. In this paper, we address this limitation and define an approximate metric space for the manifold of diffeomorphisms $\mathcal{G}$. We propose a method to break down the large deformation into finite compositions of small deformations. This results in a broken geodesic path on $\mathcal{G}$ and its length now forms an approximate registration metric. We illustrate the method using a simple, intensity-based, log-demon implementation. Validation results of the proposed method show that it can capture large and complex deformations while producing qualitatively better results than the state-of-the-art methods. The results also demonstrate that the proposed registration metric is a good indicator of the degree of deformation.
\end{abstract}


\end{frontmatter}


\tikz[overlay,remember picture]
{
    \node at ($(current page.west)+(20.5,0)$) [rotate=90] { This paper is under consideration at Computer Vision and Image Understanding};
}
\section{Introduction}
Computational anatomy is an area of research focused on developing computational models of biological organs to study the anatomical variabilities in the deformation space. Anatomical variations arise due to structural differences across individuals and changes due to growth or atrophy in an individual. A common approach for quantifying the variability is to register the different images/shapes and parameterize the deformations with velocity fields.
The registration algorithms typically optimize an energy functional based on a similarity function computed between the fixed and moving images.
\\
Many initial image registration attempts use energy functionals inspired by physical processes to model the deformation as an elastic deformation  \cite{phys1}, or viscous flow  \cite{phys2} or diffusion   \cite{phys3}. The diffusion-based approaches have been explored for 3D medical images in general \cite{understand_demons} and with deformations constrained to be diffeomorphic \cite{logdemon} to ensure preservation of the topology.
The two main approaches used to capture diffeomorphisms are parametric and nonparametric methods. The Free Form Deformation (FFD) model \cite{FFD1,FFD} is a widely used parametric deformation model for medical image registration, where a rectangular grid with control points is used to model the deformation. Large diffeomorphic deformations \cite{FFD} are handled by concatenating multiple FFDs. Deformable Registration via Attribute Matching and Mutual-Saliency Weighting (DRAMMS) \cite{dramms} is a popular FFD-based method, which also handles inter-subject registration. DRAMMS matches Gabor features and prioritizes the reliable matching between images while performing registration. The main drawback of the deformations captured by FFD models is that they do not guarantee invertibility. The non-parametric methods represent the deformation with stationary or time varying velocity vector field. The diffeomorphic log-demon \cite{logdemon} is an example of the former while the Large Deformation Diffeomorphic Metric Mapping (LDDMM) \cite{lddmm2} inspired from \cite{lddmmb} is an example of the latter approach. In LDDMM, deformations are defined as geodesics on a Riemannian manifold, which is attractive; however, the methods based on this framework are computationally complex. The diffeomorphic log-demon framework \cite{logdemon}, on the other hand,  assigns a Lie group structure and assumes a stationary velocity field (SVF) which leads to computationally efficient methods, which is of interest to the community for practical purposes. This has motivated the exploration of a stationary LDDMM framework \cite{lddmmsvf} that leverages the SVF advantage. The captured deformations are constrained to be symmetric in time-varying LDDMM \cite{ants} and log-demon \cite{symlogdemon} methods. The log-demon framework is of interest to the community for practical purposes because of its computational efficiency and simplicity.

The Lie group stucture gives a locally defined group exponential map to map the SVF to the deformation. Thus log-demon framework is meant to capture only neighboring elements in the manifold, i.e., only a limited degree of deformations can be captured. This will be referred to as the limited coverage issue of the SVF methods in this paper.

Notwithstanding the limited coverage, several SVF based methods have been reported for efficient medical image registration with different similarity metrics, sim, such as local correlation between the images \cite{SVF_variants2}, spectral features \cite{SVF_variants1}, modality independent
neighborhood descriptors \cite{SVF_variants3} and wavelet features \cite{SVF_variants4,SVF_variants5}.  

Lie group exponential map need not have any Riemannian metric associated with it as it is derived from the Lie group properties, and the log-demon framework is known to be metric free. The framework however, helps to analyze the underlying manifold and develop a representative group element called a template to study the variabilities among the manifold elements. The log-Euclidean mean \cite{mean1} and the bi-invariant mean \cite{mean2} are commonly used alternative approaches in diffeomorphic demons-based methods for template creation. Defining a metric space with SVF methods embedding complex deformations, is a challenging problem. A shape metric approximation method has been explored in \cite{Yang2015} for the Lie group exponential map, with the shape metric taken to be the minimal length group exponential path connecting two shapes.

Defining a metric space with broader coverage for diffeomorphic deformations with Lie group structure is still unaddressed in the literature. The large deformations of interest are those that arise in the registration of inter-subject images/shapes. In this paper, we propose an image registration framework which can capture large diffeomorphisms. Any of the SVF based registration algorithms can be used in this framework. SVF based algorithms cannot handle complex deformations because the deformations are constrained to be smooth for the entire image and thus constrain the possible degree of deformation to some extent. We address this drawback by splitting the large deformation as compositions of smaller deformations.

\section{Background}

Let $G$ be a finite-dimensional Lie group with Lie algebra $\mathfrak{g}$. Recall that $\mathfrak{g}$ is the tangent space $T_eG$ at the identity $e$ of $G$. The exponential map $\exp:\mathfrak{g} \rightarrow G$ is defined as follows: Let $v\in \mathfrak{g}$. Then $\exp(v)=\g_{v}(1)$, where $\g$ is the unique one-parameter subgroup of the Lie group $G$ with $v$ being its tangent vector at $e$. The vector $v$ is called the infinitesimal generator of $\g$. 
  The exponential map is a diffeomorphism from a small neighborhood containing 0 in the Lie algebra $\mathfrak{g}$ to a small neighborhood containing $e$ of $G$. 
  
  Due to the fact that a bi-invariant metric may not exist for most of the Lie groups considered in medical image registration, the deformations considered here are elements of a Lie group with the Cartan-Schouten Connection \cite{ccs}. This is the same as the one considered in the log-demon framework \cite{logdemon}.
  This is a left invariant connection \cite{mean1} in which geodesics through the identity are one-parameter subgroups. The group geodesics are the geodesics of the connection. Any two neighboring points can be connected with a group geodesic. That is, if the stationary velocity field $v$ connecting two images in the manifold $\mathcal{G}$ is small enough, then its group exponential map forms a geodesic. Similarly every $\mathfrak{g} \in G$ has a geodesically convex open neighbourhood \cite{mean1}.

\section{Proposed method}
SVF based registration methods capture only a limited degree of deformation because  exponential mappings are only locally defined. In order to perform registration of a moving image towards a fixed image, SVF is computed iteratively by updating it with a smoothed velocity field. This update is computed via a similarity metric that measures the correspondence between the moving and fixed images. The spatial smoothing has a detrimental effect as we explain next. A complex deformation typically consists of spatially independent deformations in a local neighbourhood. Depending on the smoothing parameter value, only major SVF updates in each region is considered for registration. Thus, modeling complex deformations with a smooth stationary velocity field is highly dependent on the similarity metric and the smoothing parameter in a registration algorithm. Finding an ideal similarity metric and an appropriate smoothing parameter applicable for any registration problem, irrespective of the complexity of the deformation and the type of data, is difficult. We propose to address this issue as follows: Deform the moving image toward the fixed image by sequentially applying an SVF based registration. The SVF based algorithm chooses the major or the predominant (correspondence-based) deformation component among the spatially independent deformations in all the neighbourhoods to register along these predominant directions. The subsequent steps in the algorithm captures the next set of predominant directions sequentially. These sequentially captured deformations has a decreasing order of degree of pixel displacement caused by the deformations. Mathematically speaking, the discussion above can be summarised as follows. Consider complex deformations as a composition of finite group geodesics and use a registration metric approximation to quantify the deformation between two images in terms of the length of a broken geodesic connecting them; a broken geodesic is a piecewise smooth curve, where each curve segment is a geodesic.

 In the proposed method, the similarity-based metric selects the predominant here deformation in each sequential step. The deformation that can bring the moving image in a step maximally closer to the target is selected from the one-parameter subgroup of deformations. In the  manifold $\mathcal{G}$ every geodesic is contained in a unique maximal geodesic. Hence the maximal group geodesic $\g_i$ computed using log-demon registration framework deforms the sequential image  $S_{i-1}$  in the previous step maximally closer to $S_N$.  The maximal group geodesic paths are composed to get the broken geodesic path.  As the deformation segments are diffeomorphic, the composed large deformation of the segments is also diffeomorphic.

\begin{figure}
\centering
\includegraphics[scale=.6]{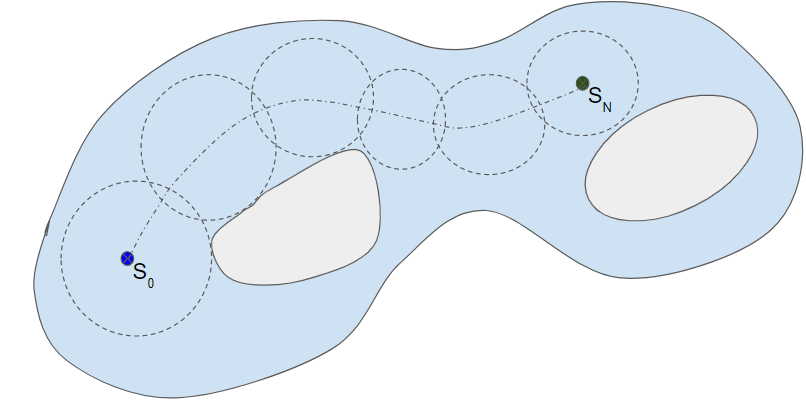}
\caption{Underlying manifold structure and a broken geodesic path connecting $S_0$ and $S_N$.}
\label{fig1.2}
\end{figure}

 In the proposed method, the coverage of the SVF method and the degree of deformation determines the number of subgroups $N$ needed to cover the space. The feature based SVF methods in general, give more coverage for a single such subgroup and reduce the value of $N$. The local deformations on the manifold $\mathcal{G}$ form a one-parameter subgroup, which is a  finite-dimensional path connected subgroup. The topological space formed by composing the group geodesics also forms a path connected manifold, as shown in Figure \ref{fig1.2}.  So a broken geodesic is a good choice to capture complex deformations as it can be computed between any two points in the manifold along with the metric associated with the path.
\\A broken geodesic $\gamma:[0,T]\rightarrow M$ has finite number of geodesic segments $\g_i$ for partitions of the domain $0<t_1< t_2<\cdots < t_{i}<\cdots t_{N}= T$  where $i=1,....N$. 
  The proposed algorithm to deform $S_0$ towards $S_N$  is given in  Algorithm \ref{g1}.
  We have chosen the registration algorithm from \cite{symlogdemon}  to compute SVF, $u_i$, in  Algorithm 1. The Energy term is as defined below where the first term is a functional of the similarity measure,
which captures the correspondence between images, with $\mbox{sim}(S_N,S_{i})=S_N-S_{i-1} \circ \exp(v_i)$. The second term is a regularization term, with Reg$(\g_{i})=\lVert \bigtriangledown \g_{i}\rVert^2$.
 \begin{equation}
     \mbox{Energy}(S_i,S_N)=\mbox{sim}(S_N ,S_i)+\mbox{Reg}(\g_i).
     \label{eq1}
 \end{equation}

\begin{algorithm}[H]
  \caption{Proposed Algorithm}
\begin{algorithmic}[1]
\State Input: $S_0$ and $S_N$
\State Result: Transformation $\g=\exp(v_1)\circ \exp(v_2)\circ ...\exp(v_{N})$
\State Initialization: $E_{\mathrm{min}}$=Energy$(S_0,S_N$) 
\Repeat
  \State  Register $S_{i-1}$ to $S_N$ $\rightarrow$  $u_i$ 
  \State $\mbox{Temp}= S_{i-1} \circ \exp(u_i)$
  \State $E_i=\mbox{Energy}(\mbox{Temp},S_N)$
  \If{$E_i<E_{\mathrm{min}}$}
   \State $v_i=u_i$
   \State $E_{\mathrm{min}}=E_i$
   \State $S_{i}= \mbox{Temp}$
  \EndIf
\Until{Convergence}
 \end{algorithmic}
 \label{g1}
 \end{algorithm}

\subsection{Registration metric approximation}

Let $\gamma$ be a broken geodesic decomposed into $N$ geodesics $\gamma_i$ with stationary field $v_i$, i.e. $\dot{\gamma_i}=v_{i}(\g(t))\in T_{\g_{i}(t)}M$.
Each of the constant velocity paths $\gamma_i$ is parameterized by the time interval $[t_{i-1},t_i]$, and  $N \in \mathbb{N}$ is minimized by requiring each of the geodesics in the broken geodesic to be maximal geodesics.
The length of the broken geodesic is defined as,
\begin{equation}
    l(\gamma)=\sum_{i}^{N}l(\gamma_i)=\sum_{i}^{N}d(S_{i-1},S_{i})
    \label{eqn1}
\end{equation}
where, $d$ is a distance metric defined in Equation \ref{q2} .  
\begin{equation}
    d(S_{i-1},S_i)=\inf\{\left \| v_i\right \|_V, S_{i-1}\circ \exp(v_i)=S_i\}.
    \label{q2}
\end{equation}
A registration metric needs to be defined to quantify the deformation between two images. The shape metric approximation in \cite{Yang2015} can be used for the group geodesics of the  Cartan-Schouten connection defined in the finite dimensional case as no bi-invariant metric exists. The length of a broken geodesic $l(\g)$ on the manifold $\mathcal{G}$ connecting $S_0$ and $S_N$, computed by Equation \ref{eqn1} is defined as the proposed approximate metric. If any two points in the manifold $\mathcal{G}$ can be connected with a broken geodesic,  then this would imply that the manifold is connected. Furthermore, there exists a notion of an asymmetric registration metric.

\section{RESULTS}
The proposed method was implemented using a simple intensity based diffeomorphic log-demon technique \cite{log-demon} for illustrating the concept which is openly available at: \url{http://dx.doi.org/10.17632/29ssbs4tzf.1}. This choice also facilitates understanding the key strengths of the method independently. Two state-of-the-art (SOTA) methods are considered for performance comparison with the proposed method: the symmetric LDDMM implementation in ANTs \cite{ants} and DRAMMS which is a feature based, free-form deformation estimation method \cite{dramms}. These two methods are considered to be good tools for inter-subject registration \cite{comp}. Publicly available codes were used for the SOTA methods with parameter settings as suggested in \cite{comp} for optimal performance. Both methods were implemented with B-spline interpolation, unless specified. 3D registration was done and the images used in the experiments were sourced from \cite{dataset} and \cite{ours} unless specified.  \\
\begin{figure}
\centering
\includegraphics[scale=.6]{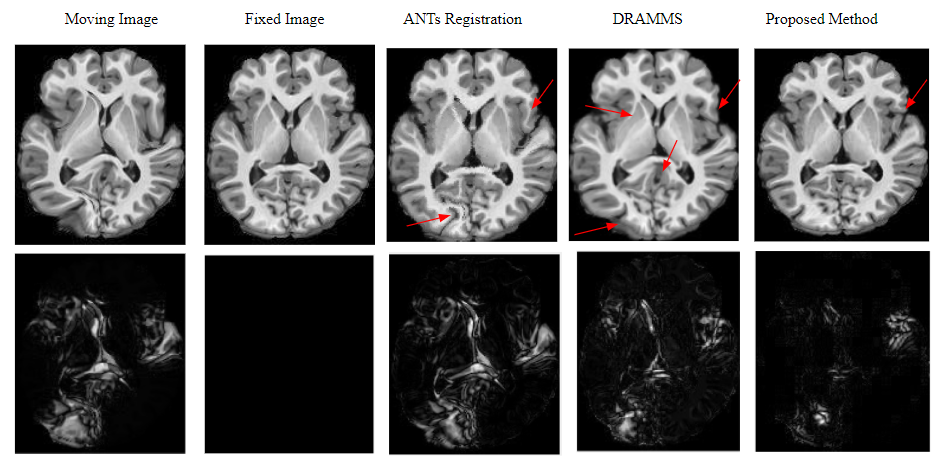}
\caption{Registration results for different methods. First row: pairs of images to be registered and registration results. Second row: absolute intensity difference between the fixed image and the rest of the images in the first row.}
\label{fig1}
\end{figure}
The proposed method and registration metric were validated with both controlled and natural deformations.
\subsection{Experiments with controlled deformations}
 An arbitrary deformation was introduced to a 2D slice of MRI and the two were registered using ANTS, DRAMMS and the proposed method. 

The results in Fig.\ref{fig1} indicate that SOTA methods have difficulty in dealing with complex deformations (regions marked with red arrows), as errors can be found in the registered results. 

\begin{figure}
\centering
\includegraphics[scale=.5]{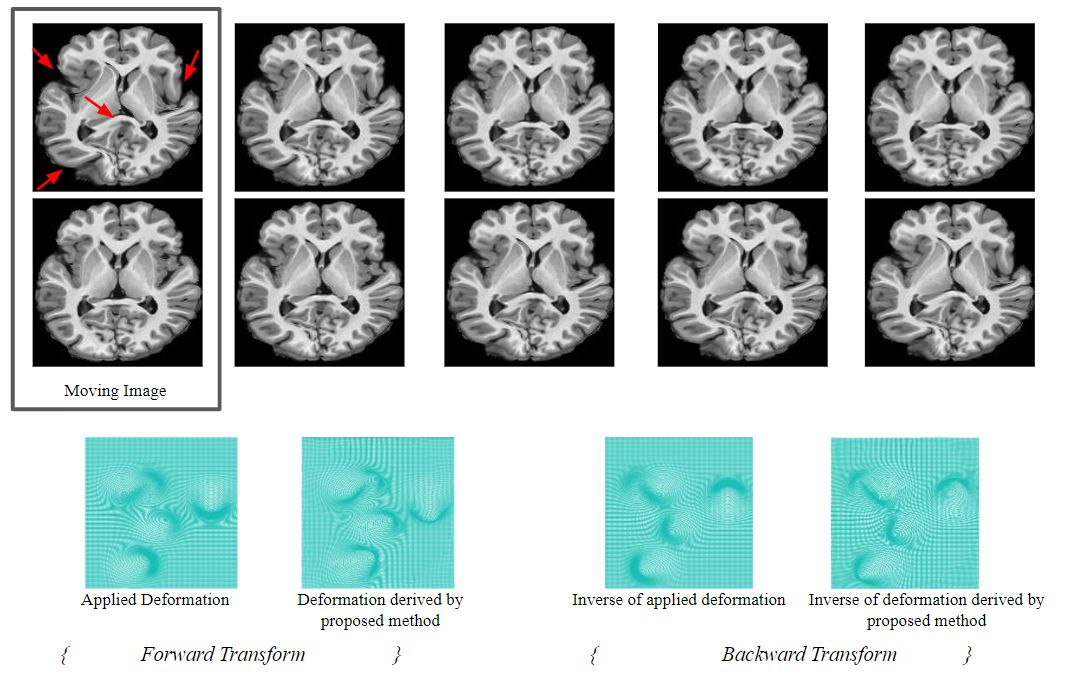}
\caption{Diffeomorphic registration with the proposed method. Geodesic forward (first row) and backward path (second row) along with the deformations captured by the proposed method.}
\label{fig2}
\end{figure}

 Next the deformation captured by proposed method was verified to be diffeomorphic. Fig.\ref{fig2} shows details of the 2D registration of the images in Fig.\ref{fig1} in terms of the forward and reverse deformations computed with a broken geodesic path. Included are the deformed images in each path. This provides a  visual confirmation that the method captures diffeomorphic forward and inverse deformations. The deformed regions are pointed by the red arrow and can also be interpreted from the applied deformation.

\begin{figure}
\centering
\includegraphics[scale=.6]{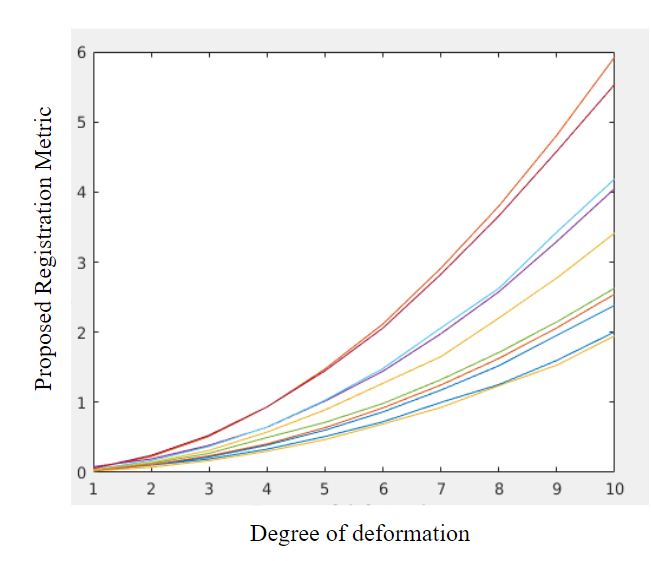}
\caption{Proposed registration metric as a function of increasing degree of deformation for 10 random deformations (the coloured plots)}
\label{graph}
\end{figure}

The proposed registration metric was validated as follows. Synthetic deformations were applied to a 2D MRI slice at 25 random control points in the image. The deformation was varied from approximately $0.05\%$ area coverage per control point in 10 equal intervals, i.e. $k\times$ $0.05\%$; $k=1,2...10$ (these are referred to as the degree of deformation). Fig.\ref{graph} shows a plot of the computed value of the proposed registration metric in Equation \eqref{eqn1}, as a function of the degree of deformations, $k$, for 10 random deformations. The monotonic behaviour seen in the plot confirms the  integrity of the proposed registration metric and the method for registration.

\subsection{Experiments with inter-subject image registration}
\begin{figure}
\centering
\includegraphics[scale=.4]{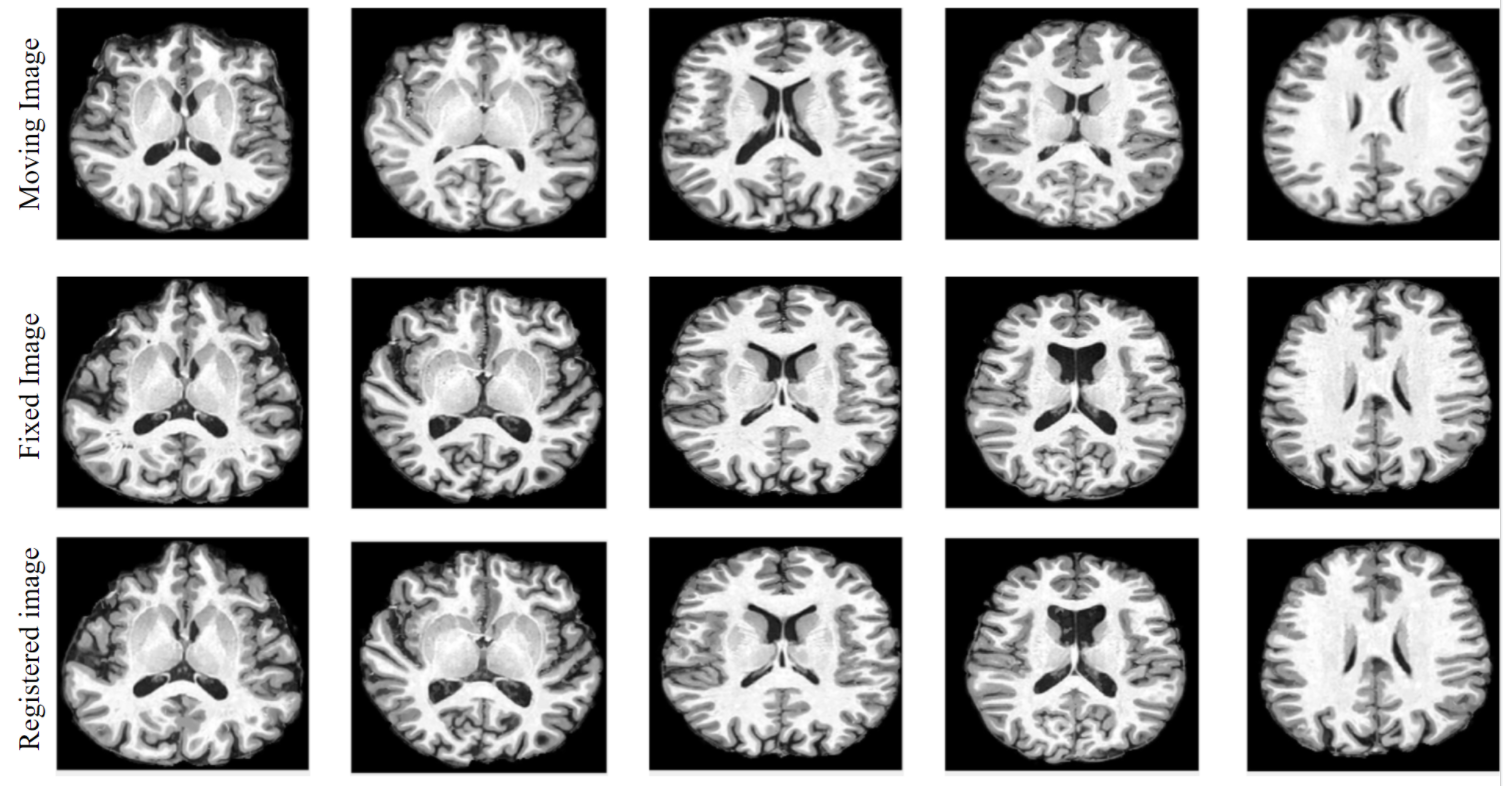}
\caption{Inter-subject image registration for 5 pairs of volumes (in 5 columns). Only sample slices are shown.}
\label{im4.5}
\end{figure}

\begin{figure}
\includegraphics[scale=.7]{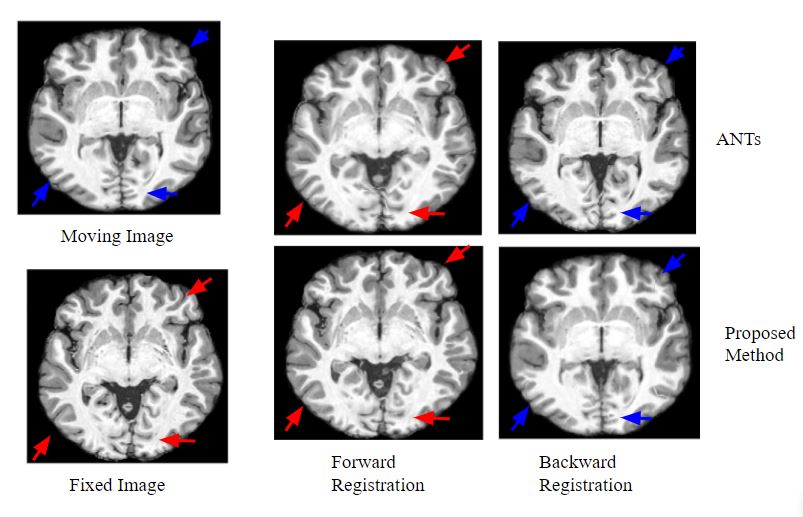}
\caption{Forward and Backward Image Registration. Blue (Red) arrow shows deformations in $S_0$ ($S_N$) and $S_N$ ($S_0$) registered to $S_0$ ($S_N$). The proposed method captures finer details compared to ANTs.}
\label{im5}
\end{figure}

\begin{figure}
\includegraphics[scale=.6]{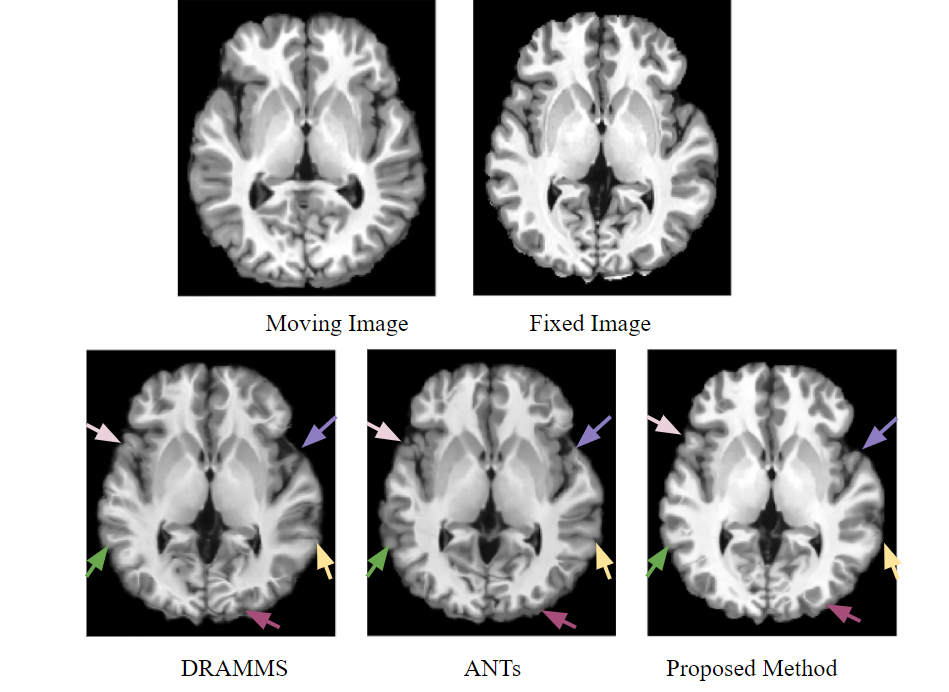}
\caption{Inter-subject image registration with 3 methods: DRAMMS, ANTs and the proposed method, implemented with linear interpolation. The regions near same colour arrows can be compared to check the registration accuracy.}
\label{im4.6}
\end{figure}

The proposed image registration algorithm was used to register MRIs of different individuals. The mean squared error (MSE) was used as a similarity metric along with cubic interpolation. To analyse the performance visually, ten 3T MRI scans were collected.  Five images collected from 20-30 year old male subjects were considered as moving images and five images collected from 40-50 year old female subjects  were considered as fixed images. Performing a good registration is challenging with this selection of moving and fixed images. The high resolution MRI scans used for this experiment are openly available at \url{http://dx.doi.org/10.17632/gnhg9n76nn.1}. The registration results for these five different pairs are shown in Fig.\ref{im4.5} where only a sample slice is visualized for the 5 cases.
The quality of registration can be assessed by observing the degree of match between images in the last two rows of each column. The results indicate that the proposed method is good at capturing complex inter-subject deformations.
The deformation calculated by the proposed method was also checked to ascertain if it is a diffeomorphism or not. Fig.\ref{im5} shows the forward and reverse deformation results with results of ANTs registration method included for comparison. The arrows overlaid on the registered images highlight regions where the proposed method yields error-free results as opposed to the other method. The forward and backward transforms in the last row compare well with the applied versions.
The performance of the proposed method on medical images to capture natural deformations was next evaluated for inter-subject image registration and compared with the state-of-the-art methods in Fig.\ref{im4.6}. To apply the computed deformation, linear interpolation was used in all the methods. ANTs and the proposed method used MSE as a similarity metric for fair comparison and DRAMMS used its Gabor feature-based metric as it is a feature based method. The results shows that the deformations at the sulcal regions are better captured by the proposed method. 
  
Next we present a quantitative comparison of the proposed method compared with ANTs and DRAMMS under the same setting. The average MSE for the 10 pair registration with ANTs was $0.0036 \pm 0.0009$, with DRAMMS it was $0.0113 \pm 0.0068$ and with the proposed method it was $0.0012 \pm 7.0552e-08$. 

\begin{figure}
\centering
\includegraphics[scale=.7]{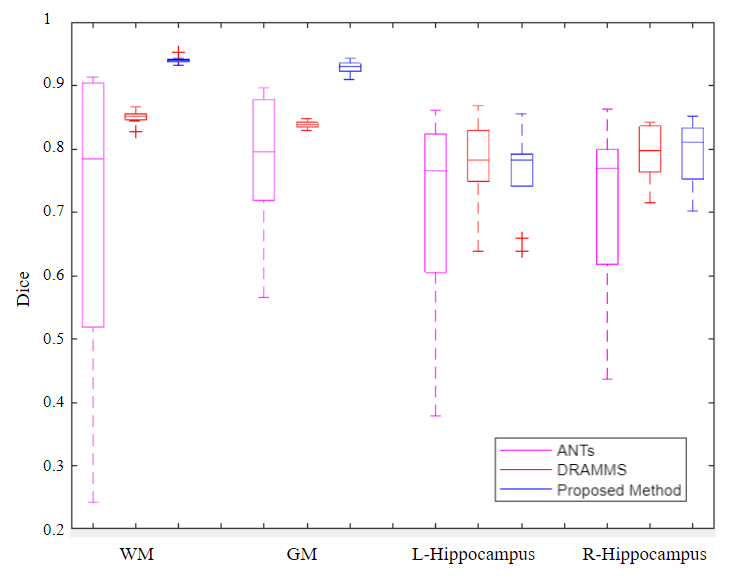}
\caption{Assessment of registration via segmentation of different structures using ANTS
(magenta), DRAMMS (red), and the proposed method (blue). Box plots for the Dice coefficient are shown for White Matter (WM), Gray Matter (GM) and the Left and Right Hippocampi. }
\label{fig_dice}
\end{figure}

The computed deformations in each method were used to transfer region segmentation (labels) from the moving image to the fixed image. The transferred segmentations are assessed using the Dice metric. Fig.\ref{fig_dice} shows a box plot of the obtained Dice values calculated by registering 10 pairs of brain MRIs with the fixed image, for white matter (WM), grey matter (GM) and 2 structures (L \& R-hippocampus). The  segmentation results  for larger structures (i.e., WM and GM) are better with the proposed method compared to the other methods, whereas the smaller structure segmentation is comparable to DRAMMS. 

These results taken together with the qualitative results, indicate that DRAMMS is not able to capture very complex deformations though the errors suffered by the label transfer is comparable to the proposed method. 

\begin{figure}
\centering
\includegraphics[scale=.6]{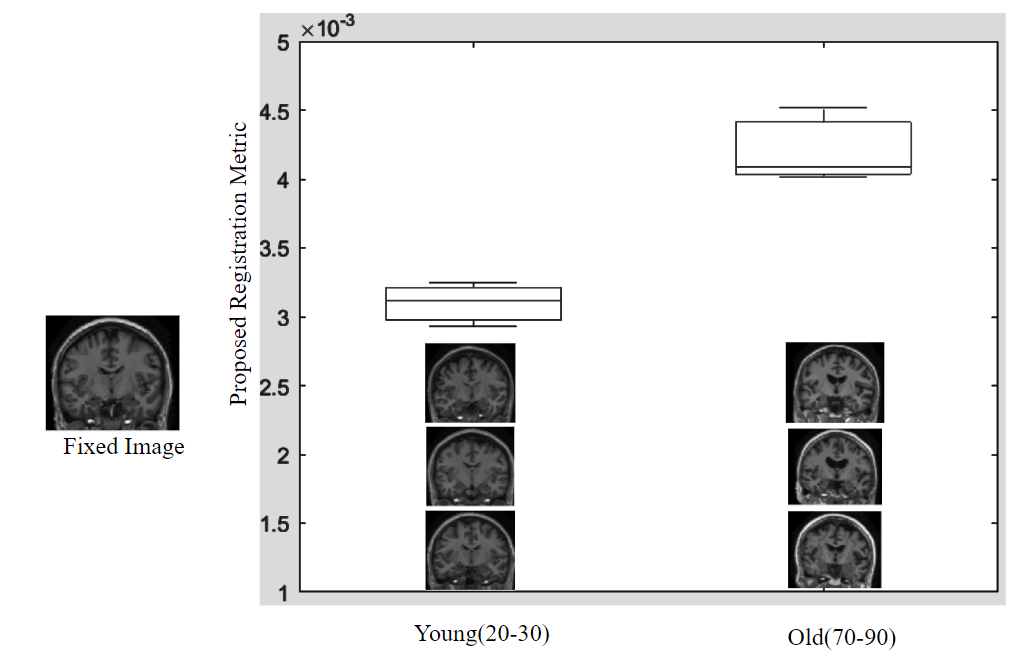}
\caption{Validation of the proposed registration metric. Box plots of the proposed registration metric values for  registration of the fixed image with images of young and old subject group.Only the central slice is displayed.}
\label{fig5}
\end{figure}

Finally, a validation of the proposed registration metric was done using two age-differentiated (20-30 versus 70-90 years) sets of MRIs, of 6 female subjects. Images from these 3D image sets were registered to an (independently drawn) MRI of a 20 year-old subject. The proposed registration metric was computed for the 6 pairs of registrations. A box plot of the registration metric value for each age group is shown in Fig.\ref{fig5}. Since the fixed image is that of a young subject, the registration metric value should be higher for the older group than for the younger group, which is confirmed by the plot. Hence, it can be concluded that the proposed registration metric is a good indicator of natural deformations as well.
\section{Conclusion}
Group exponential map based methods, with simple similarity registration metrics, fail to capture large deformations as the map is local in nature. We have addressed this issue in this paper by modelling large deformations with broken geodesic paths with the path length taken to be the associated registration metric. The proposed framework/method makes the underlying manifold path-connected. The results of implementation with a simple log-demon method show the performance to be superior to SOTA methods for complex/large deformations. Efficient implementation of the proposed approach are currently being explored.
\section*{Declaration of interest}
The authors declare that they have no known competing financial interests or personal relationships that could have appeared to influence the work reported in this paper.

\bibliography{mybibfile}

\end{document}